# Ordering-Based Search: A Simple and Effective Algorithm for Learning Bayesian Networks

**Marc Teyssier**
Computer Science Dept.
Stanford University
Stanford, CA 94305

**Daphne Koller**
Computer Science Dept.
Stanford University
Stanford, CA 94305

## Abstract

One of the basic tasks for Bayesian networks (BNs) is that of learning a network structure from data. The BN-learning problem is NP-hard, so the standard solution is heuristic search. Many approaches have been proposed for this task, but only a very small number outperform the baseline of greedy hill-climbing with tabu lists; moreover, many of the proposed algorithms are quite complex and hard to implement. In this paper, we propose a very simple and easy-to-implement method for addressing this task. Our approach is based on the well-known fact that the best network (of bounded in-degree) consistent with a given node ordering can be found very efficiently. We therefore propose a search not over the space of structures, but over the space of orderings, selecting for each ordering the best network consistent with it. This search space is much smaller, makes more global search steps, has a lower branching factor, and avoids costly acyclicity checks. We present results for this algorithm on both synthetic and real data sets, evaluating both the score of the network found and in the running time. We show that ordering-based search outperforms the standard baseline, and is competitive with recent algorithms that are much harder to implement.

## 1 Introduction

Much work has been done on the problem on learning the structure of a Bayesian network (BN) [Pearl, 1988] from a data set $\mathcal{D}$. The task can be formulated as that of finding a network structure that maximizes some scoring function defined relative to $\mathcal{D}$. Although several definitions of score have been proposed, the most commonly used is the Bayesian score [Cooper and Herskovits, 1992], and specifically its BDe variant [Heckerman et al., 1995]. Unfortunately, the task of finding a network structure that optimizes the score is a combinatorial optimization problem, and is known to be NP-hard [Chickering, 1996a; Chickering et al., 2003], even if we restrict each node to having at most two parents.

The standard methodology for addressing this problem is to perform heuristic search over some space. Many algorithms have been proposed along these lines, varying both on the formulation of the search space, and on the algorithm used to search the space. However, it has proven surprisingly hard to beat the simple yet highly effective baseline, of greedy hill-climbing search over the space of network structures, modified with a tabu list and random restarts. Recently, several more powerful algorithms have been proposed (e.g., [Chickering, 1996b; Chickering, 2002; Steck, 2000; Elidan et al., 2002; Moore and Wong, 2003]). Although some of these approaches have been shown to provide improvements over the baseline, they tend to be fairly complex and hard to implement.

In this paper, we define a simple yet effective algorithm for finding a high-scoring network structure. Our approach is based on the fundamental observation [Buntine, 1991; Cooper and Herskovits, 1992] that, given an ordering $\prec$ on the variables in the network, finding the highest-scoring network consistent with $\prec$ is not NP-hard. Indeed, if we bound the in-degree of a node to $k$, this task can be accomplished in time $O(n^k)$ (where $n$ is the number of variables). By itself, this observation is of limited use, as determining an appropriate ordering is itself a difficult problem, usually requiring significant domain knowledge.

However, this observation leads to the obvious idea of conducting our search over the space of orderings $\prec$, rather than specific network structures. We define the score of an ordering as the score of the best network consistent with it. We define a set of local search operators — flipping a pair of adjacent nodes in the ordering — that traverse the space of orderings, and use greedy hill-climbing search, with a tabu list and ran-dom restarts, to find an ordering that maximizes the score. The idea of using orderings to provide a more global view on structures was also used by Larranaga *et al.* [1996], who proposed an approach similar to ours, but using a genetic algorithm search over structures. Their algorithm is quite complex, and its advantages in practice are unclear. In a different setting, Friedman and Koller [2003] used MCMC over the space of orderings for Bayesian model averaging for structure discovery.

The ordering search space turns out to have some useful properties. First, it is significantly smaller than the space of network structures: $2^{\Omega(n^2)}$ network structures versus $2^{O(n \log n)}$ orderings. Second, each step in the search makes a more global modification to the current hypothesis, thereby better avoiding local minima. Third, the branching factor in our search space is $O(n)$ rather than $O(n^2)$, reducing the cost of evaluating candidate successors in each step. Finally, as acyclicity is not an issue given an ordering, we avoid the need to perform acyclicity checks on candidate successors, a potentially costly operation for large networks.

The main disadvantage of ordering-based search is the need to compute, in advance, a large set of sufficient statistics: for each variable and each possible parent set. This cost can be particularly high if the number of data instances is large. We reduce this cost by using both the AD-tree data structure of Moore and Lee [1997], and by pruning the space of possible parents for each node using the method of Friedman *et al.* [1999]. We note that, when the data set is very large, we can also reduce the computational burden by learning from a randomly-sampled subset.

We experimented with ordering-based search, comparing it to the standard baseline of greedy-tabu search over structures. We used both synthetic data, generated from a known network, and real data in the domain of gene expression. We show that, for domains with a large number of variables, our method is less likely to get trapped in local minima, and therefore often finds significantly higher-scoring structures. Moreover, it generally finds its final solution much faster than structure-based search (including the required pre-processing time). We also performed a partial comparison to the (extensive) experimental results of Moore and Wong [2003], showing that our simpler method finds networks of comparable quality, and often using less computation. Overall, our results suggest that this very simple and somewhat obvious method is a surprisingly effective approach for model selection in Bayesian network learning.

## 2 BN Structure Learning

Consider the problem of analyzing the distribution over some set $\mathcal{X}$ of random variables $X_1, \ldots, X_n$, each of which takes values in some domain $Val(X_i)$. For simplicity, we focus on the case where the variables are discrete-valued, but our approach extends easily to the continuous case. Our input is a fully observed data set $\mathcal{D} = \{\boldsymbol{x}[1], \ldots, \boldsymbol{x}[M]\}$, where each $\boldsymbol{x}[m]$ is a complete assignment to the variables $X_1, \ldots, X_n$ in $Val(X_1, \ldots, X_n)$. Our goal is to find a network structure $G$ that is a good predictor for the data.

The most common approach to this task is to define it as an optimization problem. We define a *scoring function score*$(G : \mathcal{D})$, which evaluates different networks relative to the data $\mathcal{D}$. We then need to solve the combinatorial optimization problem of finding the network that achieves the highest score. For the remainder of this discussion, we take the training set $\mathcal{D}$ to be fixed, and omit mentioning it explicitly.

Several scoring functions have been proposed; most common are the BIC/MDL score [Schwarz, 1978] and the BDe score [Heckerman *et al.*, 1995]. The details of these scores are not relevant for our discussion. For our purpose, we assume only two properties, shared by these scores and all others in common use. First, that the score is *decomposable*, i.e., that it is the sum of scores associated with individual families (where a family is a node and its parents):

$$score(G) = \sum_{i=1}^{n} score(X_i, \text{Pa}_G(X_i)).$$

The second property is the reduction of the score to *sufficient statistics* associated with individual families. In the discrete case, these statistics are simply frequency counts of instantiations within each family: $M[x_i, \boldsymbol{u}]$ for each $x_i \in Val(X_i), \boldsymbol{u} \in Val(\text{Pa}_G(X_i))$.

Given a scoring function, our task is finding

$$\text{argmax}_G score(G) \qquad (1)$$

This task is a hard combinatorial problem. Several of its specific instantiations have been shown to be NP-hard, even when the maximum number of parents per node is at most two [Chickering, 1996a; Chickering *et al.*, 2003]. The key intuition behind this result is that, due to the global acyclicity constraint, the choice of parent set for one node imposes constraints on the possible parent sets for other nodes.

One method for circumventing this problem is to postulate a pre-determined ordering $\prec$ over $X_1, \ldots, X_n$, and restrict the graph $G$ to be consistent with that ordering: If $X_i \in \text{Pa}_G(X_j)$, then $X_i \prec X_j$. This constraint ensures that all such consistent structures are acyclic, rendering the choice of parent set for different nodes independent. This observation was used by most of the early algorithms for BN structure learning, which searched for a network consistent with a pre-determined ordering (e.g., [Cooper and Herskovits, 1992]). Unfortunately, coming up with a good ordering requires a significant amount of domain knowledge, which is not commonly available in many practical applications. Therefore, most recent algorithms for BN structure learning do not make this assumption, and search over the general space of network structures.

The most common solution for finding a high-scoring network is some variant of local search over the space of networks using the operators of edge addition, deletion, and reversal. The decomposability property and the use of sufficient statistics allow these operators to be evaluated very efficiently. Most typically, the algorithm performs greedy hill-climbing search, with occasional random restarts to address the problem of local maxima. An important improvement that also avoids local maxima is the use of a *tabu list*, that prevents the algorithm from undoing operators (such as an edge addition) that were performed only recently. Despite the simplicity of this approach, and despite extensive attempts to find better methods, this baseline has proven hard to beat.

## 3 Search over Orderings

### 3.1 Search Space

In this paper, we propose a very simple approach, based on the observation that finding the best network consistent with a given ordering $\prec$ is much easier than the general case. We restrict the network in-degree — the number of parents per node — to a fixed bound $k$. We note that this assumption is commonly used in BN structure learning algorithms, to help reduce the fragmentation of the data and the resulting over-fitting to the training set. For a given ordering $\prec$ we can now define the possible parent sets for the node $X_i$:

$$\mathcal{U}_{i,\prec} = \{\boldsymbol{U} \;:\; \boldsymbol{U} \prec X_i, |\boldsymbol{U}| \leq k\}. \qquad (2)$$

where $\boldsymbol{U} \prec X_i$ is defined to hold when all nodes in $\boldsymbol{U}$ precede $X_i$ in $\prec$. The number of such parent sets is at most $\binom{n}{k}$. The optimal parent set for each node $X_i$ is simply

$$\mathrm{Pa}_\prec(X_i) = \mathrm{argmax}_{\boldsymbol{U} \in \mathcal{U}_{i,\prec}} score(X_i, \boldsymbol{U}) \qquad (3)$$

We can thus find the optimal parent set for each node in time $O(nf_{\max})$, where $f_{\max}$ is the maximal number of possible families per node. In this formulation, $f_{\max} = \binom{n}{k} = O(n^k)$. As the decisions for different nodes do not constrain each other, this set of selected families provides the optimal network $G^*_\prec$ consistent with $\prec$ and the in-degree bound $k$. We define this network to be $G^*_\prec$.

In the unconstrained case, we note that any acyclic (directed) graph is consistent with some ordering. Hence, the optimal network with no ordering constraint is simply the network $G^*_{\prec^*}$ for

$$\prec^* = \mathrm{argmax}_\prec score(G^*_\prec).$$

We can therefore find the optimal network by finding the optimal ordering, where the score of an ordering is the score of the best network consistent with it.

We perform this search using the same simple yet successful approach of greedy local hill-climbing with random restarts and a tabu list. Our state space is the set of ordering $\mathcal{O}$. While there are several possible sets of operators over this space, a simple one that worked well in our experiments is a simple swap operator:

$$(X_{i_1}, \ldots, X_{i_j}, X_{i_{j+1}}, \ldots) \mapsto (X_{i_1}, \ldots, X_{i_{j+1}}, X_{i_j}, \ldots) \qquad (4)$$

We perform the search by considering all $n-1$ candidate successors of the current ordering. We compare the delta-scores of the successor orderings obtained by these swaps — the difference between their score and the current one, and take the one that gives the highest delta-score. The tabu list is used to prevent the algorithm from reversing a swap that was executed recently in the search. We continue this process until a local maximum is reached.

### 3.2 Caching and Pruning

Importantly, as is the case for Bayesian network search, we can gain considerable savings by caching computations. Consider the operator in Eq. (4), which takes an ordering $\prec$ to another $\prec'$. The possible parent sets for a variable $X_l$ other than $X_{i_j}$ and $X_{i_{j+1}}$ remain unchanged, as the set of variables preceding them is the same in $\prec$ and $\prec'$. Thus, for each operator, we need only recompute the optimal parent sets for two variables rather than $n$.

Moreover, if we take the step of going to $\prec'$, the only "new" operators are swapping $X_{i_{j-1}}, X_{i_{j+1}}$ and swapping $X_{i_j}, X_{i_{j+2}}$. All other operators are the same in $\prec'$ and in $\prec$, and, moreover, their delta-score relative to these two search states does not change. Thus, if we move from $\prec$ to $\prec'$, we need only re-evaluate the delta-score of the two newly-created operators.

Overall, if we have computed the delta-scores for (all of the successors of) $\prec$, the cost of evaluating the delta-scores for its chosen successor $\prec'$ is only $4f_{\max}$.

Each of our steps in our search requires time which is linear in $f_{\max}$. In addition, to initialize the search we must initially compute the score, and hence the sufficient statistics, for every possible family of every variable given the initial ordering $\prec$. This cost can be quite prohibitive for large networks and/or moderate values of $k$, especially for data sets with a large number of records.

We reduce this cost in several ways. First, we use the efficient AD-tree data structure of Moore and Lee [1997] to pre-compute the sufficient statistics for all of the relevant families in advance. This structure allows us to compute the sufficient statistics even for large data sets, as in practice the computation time will no longer be linear in the number of records of the data set. We then rank the possible families for each node $X_i$ according to their score, computed using the sufficient statistics. We can now find the best scoring family of $X_i$ consistent with $\prec$ by searching through this list, selecting the first family consistent with $\prec$.

We can further reduce the cost by pruning the set of possible parent sets for each node, thereby reducing the cost of each step from $f_{\max}$ to $f_{\text{eff}} < f_{\max}$. Specifically, if $\boldsymbol{U}' \subset \boldsymbol{U}$, and $score(X_i, \boldsymbol{U}') \geq score(X_i, \boldsymbol{U})$, then we can eliminate the family $\boldsymbol{U}$ from consideration: for any ordering $\prec$, if $\boldsymbol{U}$ is a legal choice as the parent set of $X_i$, so is $\boldsymbol{U}'$, so the algorithm can always pick $\boldsymbol{U}'$ over $\boldsymbol{U}$. This pruning has significant effect in practice; for example, for the ICU-Alarm network [Beinlich et al., 1989], we have observed cases where $f_{\max} = 58905$ whereas $f_{\text{eff}} \approx 250$. This pruning procedure is sound, in that it is guaranteed never to remove the optimal parent set from consideration. However, it only reduces the costs incurred during the search itself; initially, we must still pre-compute the scores of all possible parent sets.

We can reduce both costs using a heuristic pruning procedure, which is not necessarily sound, based on the *sparse candidate algorithm* of Friedman et al. [1999]. This heuristic reduces the search space by pre-selecting a small set of *candidate parents* for each variable $X_i$. We then only consider parent sets selected from among this candidate set. In our implementation, we select for each variable $X_i$ a fixed-size set of candidate parents that are the variables most strongly correlated with $X_i$. This pre-selection reduces both the number of families that we must score, and (subsequently) the number of possible parent sets considered during the search.

## 4 Experimental Results

### 4.1 Experimental Setup

We evaluated our algorithm — ordering-search — on a variety of data sets, both synthetic and real (see Table 1). We generated synthetic data sets from two Bayesian networks: the ICU-Alarm network [Beinlich et al., 1989] (*alarm1*, *alarm2*, and *alarm3*, which vary in their size) and the Diabetes network [Andreassen et al., 1991] (*diabetes*). We also used discretized version of two real world yeast gene expression data sets: one a large subset of a stress response data set [Gasch et al., 2000] (*stress*), and the other a small subset of the Rosetta yeast knockout data [Hughes et al., 2000] (*rosetta*). We also ran on three of the data sets used by Moore and Wong [2003] (*alarm4*, *letters*, and *edsgc*), allowing a direct comparison to their results. To obtain a baseline performance, we implemented the standard benchmark algorithm of greedy hill-climbing over the space of graphs, with random restarts and a tabu list (DAG-search). We use the same implementation of AD-trees for computing the sufficient statistics, and the same set of candidate parents per node, as selected by the sparse candidate algorithm. We optimized this baseline search as much as possible, and compared it with other state-of-the-art systems, so that we are convinced it is a fair benchmark.

Both search algorithms have a set of parameters that affect its performance. We selected the size of the tabu list and the number of random moves between each restart using a systematic search, picking the best set of parameters for each algorithm. The starting point of both algorithms is also chosen according to the performance of the search: for the DAG-search the best results were given starting from an empty network whereas for the ordering-search we chose a random starting point. The number of moves without improvement before restarting was selected to be the same as the size of the tabu list. The number of candidate parents per node was chosen between 10 and 30, depending on the size of the data set and on the number of nodes in the original network. For data generated from a network, we selected the maximal in-degree $k$ to be the actual maximum in-degree in the network. For real data, we set $k = 3$. These bounds were applied both to DAG-search and to ordering-search. Finally, in all runs we used the BDe score, with a uniform prior and an equivalent sample size of 5. All experiments were performed on a 700 MHz Pentium III with 2 gigabytes of RAM.

### 4.2 Results

To compare the two searches, we recorded the best scoring network found by each one, and the computation time at which this network was found. We ran each experiment at least 4 times and averaged the results. Graphical results can be found in Figure 1 and precise numerical values in Table 2.

In terms of computation time, DAG-search always produces its first result before ordering-search. This is due to the initial cost of precomputing all of the family scores. However, as soon as this computation is over, our ordering-search converges almost instantaneously. Overall, it appears that ordering-search mostly converges before DAG-search, and (in 1–2 cases) a small constant factor afterwards. In terms of maximum score obtained by the algorithm, ordering-search performs at least as well or better than DAG-search. There are two main factors which affect the difficulty of the search. First, the number of variables determines the size of the search space: the more variables, the harder the search. The second main factor is the number of records: if there are not many records in the data set, the statistical signal is fainter and it is hard for both algorithms to find the "right direction" in the search.

Table 1: Datasets used

| DATA SET | TYPE | #INSTANCES | #VARIABLES | AVG #VALS/VAR | MAX PARENTS |
|---|---|---|---|---|---|
| *alarm1* | Synthetic | 100 | 37 | 2.8 | 4 |
| *alarm2* | Synthetic | 1000 | 37 | 2.8 | 4 |
| *alarm3* | Synthetic | 10K | 37 | 2.8 | 4 |
| *alarm4* | Synthetic | 20K | 37 | 2.8 | 4 |
| *diabetes* | Synthetic | 10K | 413 | 11.3 | 2 |
| *stress* | Real life | 173 | 6100 | 3.0 | — |
| *rosetta* | Real life | 284 | 37 | 3.0 | — |
| *letters* | Real life | 20K | 17 | 3.4 | — |
| *edsgc* | Real life | 300K | 24 | 2.0 | — |

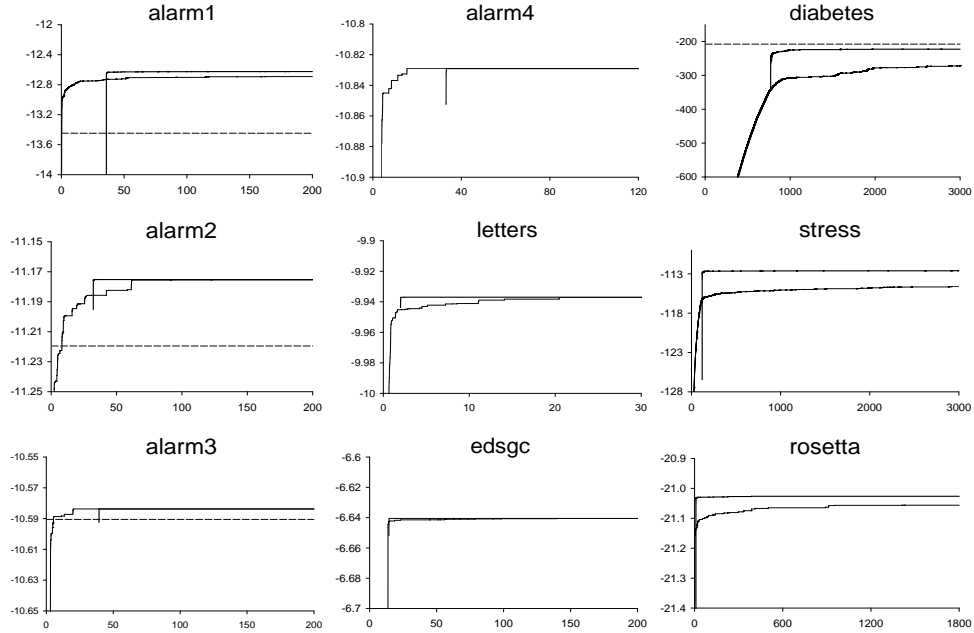

Figure 1: BDe score per datapoint versus computation time. Ordering-search is represented by the thick line and DAG-search by the thin line. When it is known, the score of the original network is represented by a dashed line. A minimal vertical axis scale of 0.1 was used except where the difference between the two searches is too large. Time is in seconds.

Table 2: Comparison of results achieved by ordering search (first number in each pair) and DAG search (second number in each pair). The best result is in bold. BDe score and log-likelihood on test data are given per datapoint. Time for convergence is in seconds.

| DATA SET | BDe SCORE | LOG-LIKELIHOOD | TIME FOR CONVERGENCE |
|---|---|---|---|
| *alarm1* | $-\mathbf{12.62}/-\mathbf{12.69}$ | $-\mathbf{9.42}/-9.47$ | $\mathbf{60}/150$ |
| *alarm2* | $-\mathbf{11.18}/-\mathbf{11.18}$ | $-\mathbf{10.34}/-\mathbf{10.34}$ | $\mathbf{33}/150$ |
| *alarm3* | $-\mathbf{10.58}/-\mathbf{10.58}$ | $-\mathbf{10.38}/-\mathbf{10.38}$ | $40/\mathbf{20}$ |
| *alarm4* | $-\mathbf{10.83}/-\mathbf{10.83}$ | $-\mathbf{10.69}/-\mathbf{10.69}$ | $35/\mathbf{16}$ |
| *diabetes* | $-\mathbf{223}/-271$ | $-\mathbf{201}/-247$ | $\mathbf{2000}/>3000$ |
| *stress* | $-\mathbf{112.6}/-114.6$ | $-\mathbf{75.8}/-76.8$ | $\mathbf{300}/>3000$ |
| *rosetta* | $-\mathbf{21.02}/-21.06$ | $-\mathbf{17.12}/-17.20$ | $\mathbf{500}/2500$ |
| *letters* | $-\mathbf{9.94}/-\mathbf{9.94}$ | $-\mathbf{9.60}/-\mathbf{9.60}$ | $\mathbf{2}/25$ |
| *edsgc* | $-\mathbf{6.64}/-\mathbf{6.64}$ | $-\mathbf{6.64}/-\mathbf{6.64}$ | $\mathbf{20}/150$ |

The results from the experiments show us that when there is roughly fewer than 50 variables and more than 1,000 records (*alarm2*, *alarm3*, *alarm4*, *letters* and *edsgc*), both algorithms find the exact same best value for each of the 4 runs of each experiment. Although it can not be determined with certainty (due to the NP-hardness of the problem), it appears likely that both algorithms find the optimal networks. When the number of variables gets larger (*diabetes*), we see that ordering-search outperforms DAG-search. We believe that this improved performance is due both to the reduced size of the ordering search space, and to the fact that ordering-search takes much larger steps in the search space, avoiding many local maxima. With small data sets (*alarm1* and *rosetta*), ordering-search also gives better results. We believe that ordering-search gains both from the fact that it takes larger steps in the search, and from the fact that the search steps themselves are faster. As the size of the data sets increases (*alarm1*, *alarm2*, *alarm3* and *alarm4*), the cost of precomputing the family scores increases whereas the statistical signal gets stronger, so that searching greedily tends to work well. As a consequence, the difference in performance between the two algorithms decreases. For the largest data set (*alarm4*), we note that the DAG-search outperforms ordering-search in terms of time for convergence (but not in terms of best scoring network found) because of the cost of preprocessing step. Finally, for the *stress* data set which has the largest number of variables (6,100) and the smallest number of instances (173), as expected, the search in the space of orderings is much more efficient.

To conclude, when the search is easy (few variables and lots of records), both algorithms find the exact same score and the solution found should be the optimum. But, when the search gets harder, DAG-search is no longer able to find the optimal solution and ordering-search finds a better scoring network.

Both search algorithms are trying to optimize the score, and hence measuring the score of the best network found is the "fair" evaluation. However, we also tested whether the differences in score translate to better generalization performance on unseen data. We therefore measured the log-likelihood of each model found on a disjoint test set generated from the same distribution. For synthetic data, we simply generated more data from the network. For real data, we used 5-fold cross-validation and averaged the results of the five folds. The results can be found in Table 2. We obtain very similar results to the experiment on BDe score. For big networks (> 50 nodes) or for small data sets (< 100), the log-likelihood of the search in the space of orderings is higher than in the space of directed graphs. This result, while satisfying, is not surprising, as BDe is known to be a reasonable surrogate for generalization performance.

Finally, we compare our results on the data sets *alarm4*, *edsgc*, and *letter* to the published results of the OR-search (optimal reinsertion) algorithm of Moore and Wong [2003]. It is hard to compare the BDe scores obtained by the two algorithms, as Moore and Wong do not report the BDe prior used in their experiments (a choice which has an effect on the score). Nevertheless, both algorithms seem to give similar best scoring networks, so it seems that they do not differ much on this point. In terms of computation time, ordering-search appears much faster, as shown in Table 3. We note that the computation time reported for OR-search is *wall clock* time whereas we use a *CPU-time* measure for ordering-search. Although these measures are not the same, this difference cannot account by itself for the fact that ordering-search is an order of magnitude faster. This conclusion is all the more safe as the results for OR-search were run on a much faster machine (a 2 GHz Pentium 4, with 2 gigabytes of RAM, versus our 700 MHz Pentium III).

Table 3: Comparison of results obtained by ordering-search (first number in each pair) with the OR-search of Moore and Wong (second number). The best result is in bold. Running times are given in seconds.

| DATA SET | BDe SCORE | RUNNING TIME |
|---|---|---|
| alarm4 | **−10.83**/**−10.83** | **35**/250 |
| letters | **−9.94**/ − 10.0 | **2**/150 |
| edsgc | −6.64/**−6.62** | **15**/100 |

## 5 Discussion and Conclusion

In this paper, we describe a simple and easy-to-implement algorithm for learning Bayesian networks from data. Our method is based on searching over the space of orderings, rather than the standard space of network structures. We note that other search spaces have also been proposed, such as the space of network equivalence classes [Chickering, 1996b;

Chickering, 2002], or the space of skeletons [Steck, 2000], but these involve fairly complex operators that are computationally expensive and hard to implement. Our results show that our method outperforms the state-of-the-art methods, both in finding higher-scoring networks and in computation time.

There are several possible extensions to this work. One can use a more clever approach to prune the space of candidate families per node (e.g., [Moore and Wong, 2003]). It would also be interesting to combine our approach with other state-of-the-art BN structure learning methods, such as the data perturbation approach of Elidan *et al.* [2002]. Finally, our method can also be applied in a straightforward way to the task of structure learning with incomplete data, using the *structural EM* approach of Friedman [1997]. We use the E-step to compute the expected sufficient statistics and thereby the expected scores, and then apply ordering-search as described. However, the computation of expected sufficient statistics requires inference, and is therefore costly; it would be interesting to construct heuristics that avoid a full recomputation at every model modification step.

**Acknowledgements.** We thank Gal Elidan for useful comments and for providing us with the *rosetta* and *stress* data sets. This work was funded by DARPA's EPCA program under sub-contract to SRI International.

# References

[Andreassen *et al.*, 1991] S. Andreassen, R. Hovorka, J. Benna, K. G. Olesen, and E. R. Carson. A model-based approach to insulin adjustment. In *Proc. Third Conference on Artificial Intelligence in Medicine*, 1991.

[Beinlich *et al.*, 1989] L. Beinlich, H. Suermondt, R. Chavez, and G. Cooper. The ALARM monitoring system: A case study with two probabilistic inference techniques for belief networks. In *Proc. Second European Conference on Artificial Intelligence in Medicine*, pages 247–256. Springer Verlag, 1989.

[Buntine, 1991] W. L. Buntine. Theory refinement on Bayesian networks. In *Proc. Seventh Annual Conference on Uncertainty Artificial Intelligence*, pages 52–60, 1991.

[Chickering *et al.*, 2003] D. M. Chickering, C. Meek, and D. Heckerman. Large-sample learning of Bayesian networks is hard. In *Proc. Nineteenth Conference on Uncertainty in Artificial Intelligence*, pages 124–133, 2003.

[Chickering, 1996a] D. M. Chickering. Learning Bayesian networks is NP-complete. In *Learning from Data: Artificial Intelligence and Statistics V*. Springer Verlag, 1996.

[Chickering, 1996b] D. M. Chickering. Learning equivalence classes of Bayesian network structures. In *Proc. Twelfth Conference on Uncertainty in Artificial Intelligence (UAI '96)*, pages 150–157, 1996.

[Chickering, 2002] D. M. Chickering. Learning equivalence classes of Bayesian-network structures. *Journal of Machine Learning Research*, 2:445–498, February 2002.

[Cooper and Herskovits, 1992] G. Cooper and E. Herskovits. A Bayesian method for the induction of probabilistic networks from data. *Machine Learning*, 9:309–347, 1992.

[Elidan *et al.*, 2002] G. Elidan, M. Ninio, N. Friedman, and D. Schuurmans. Data perturbation for escaping local maxima in learning. In *Proc. National Conference on Artificial Intelligence (AAAI)*, 2002.

[Friedman and Koller, 2003] N. Friedman and D. Koller. Being Bayesian about network structure: A Bayesian approach to structure discovery in Bayesian networks. *Machine Learning*, 50:95–126, 2003.

[Friedman *et al.*, 1999] N. Friedman, I. Nachman, and D. Pe'er. Learning Bayesian network structure from massive datasets: The "sparse candidate" algorithm. In *Proc. Fifteenth Conference on Uncertainty in Artificial Intelligence*, pages 196–205, 1999.

[Friedman, 1997] N. Friedman. Learning belief networks in the presence of missing values and hidden variables. In *Proceedings of the Fourteenth International Conference on Machine Learning*, pages 125–133. Morgan Kaufmann, San Francisco, 1997.

[Gasch *et al.*, 2000] A. Gasch, P. Spellman, C. Kao, O. Carmel-Harel, M. Eisen, G. Storz, D. Botstein, and P. Brown. Genomic expression program in the response of yeast cells to environmental changes. *Mol. Bio. Cell*, 11:4241–4257, 2000.

[Heckerman *et al.*, 1995] D. Heckerman, D. Geiger, and D. M. Chickering. Learning Bayesian networks: The combination of knowledge and statistical data. *Machine Learning*, 20:197–243, 1995.

[Hughes *et al.*, 2000] T. Hughes, M. Marton, A. Jones, C. Roberts, R. Stoughton, C. Armour, H. Bennett, E. Coffey, H. Dai, Y. He, M. Kidd, A. King, M. Meyer, D. Slade, P. Lum, S. Stepaniants, D. Shoemaker, D. Gachotte, K. Chakraburtty, J. Simon, M. Bard, and S. Friend. Functional discovery via a compendium of expression profiles. *Cell*, 102(1):109–26, 2000.

[Larranaga *et al.*, 1996] P. Larranaga, C. Kuijpers, R. Murga, and Y. Yurramendi. Learning Bayesian network structures by searching for the best ordering with genetic algorithms. *IEEE Transactions on System, Man and Cybernetics*, 26(4):487–493, 1996.

[Moore and Lee, 1997] A. W. Moore and M. S. Lee. Cached sufficient statistics for efficient machine learning with large datasets. *Journal of Artificial Intelligence Research*, 8:67–91, 1997.

[Moore and Wong, 2003] A. Moore and W.-K. Wong. Optimal reinsertion: A new search operator for accelerated and more accurate Bayesian network structure learning. In *Proc. International Conference on Machine Learning*, 2003.

[Pearl, 1988] J. Pearl. *Probabilistic Reasoning in Intelligent Systems*. Morgan Kaufmann, San Mateo, California, 1988.

[Schwarz, 1978] G. Schwarz. Estimating the dimension of a model. *Annals of Stastics*, 6:461–464, 1978.

[Steck, 2000] H. Steck. On the use of skeletons when learning in bayesian networks. In *Proc. Sixteenth Conference on Uncertainty in Artificial Intelligence (UAI '00)*, pages 558–65, 2000.